\documentclass[10pt,twocolumn,letterpaper]{article}

\usepackage[pagenumbers]{cvpr} %

\usepackage{graphicx}
\usepackage{amsmath}
\usepackage{amssymb}
\usepackage{booktabs}
\usepackage{caption}
\usepackage{subcaption}
\usepackage{multirow}
\usepackage{multicol}
\usepackage{xcolor}
\usepackage{adjustbox}
\usepackage{array}
\usepackage{tablefootnote}

\newcommand{\x}{\mathbf{x}}

\newcolumntype{R}[2]{%
    >{\adjustbox{angle=#1,lap=\width-(#2)}\bgroup}%
    l%
    <{\egroup}%
}

\usepackage[pagebackref,breaklinks,colorlinks]{hyperref}

\usepackage[capitalize]{cleveref}
\crefname{section}{Sec.}{Secs.}
\Crefname{section}{Section}{Sections}
\Crefname{table}{Table}{Tables}
\crefname{table}{Tab.}{Tabs.}

\def\cvprPaperID{*****} %
\def\confName{CVPR}
\def\confYear{2023}

\begin{document}

\title{Mobile V-MoEs: Scaling Down Vision Transformers\\via Sparse Mixture-of-Experts}

\author{Erik Daxberger\thanks{Correspondence to: Erik Daxberger, \href{mailto:edaxberger@apple.com}{\tt edaxberger@apple.com}.}
\and
Floris Weers
\and
Bowen Zhang
\and
Tom Gunter
\and
Ruoming Pang
\and
Marcin Eichner
\and
Michael Emmersberger
\and
Yinfei Yang
\and
Alexander Toshev
\and
Xianzhi Du\\
\\\hspace{-155mm}Apple\vspace{-5mm}
}

\maketitle

\begin{abstract}
Sparse Mixture-of-Experts models (MoEs) have recently gained popularity due to their ability to decouple model size from inference efficiency by only activating a small subset of the model parameters for any given input token.
As such, sparse MoEs have enabled unprecedented scalability, resulting in tremendous successes across domains such as natural language processing and computer vision.
In this work, we instead explore the use of sparse MoEs to scale-down Vision Transformers (ViTs) to make them more attractive for resource-constrained vision applications.
To this end, we propose a simplified and mobile-friendly MoE design where entire images rather than individual patches are routed to the experts.
We also propose a stable MoE training procedure that uses super-class information to guide the router.
We empirically show that our sparse Mobile Vision MoEs (V-MoEs) can achieve a better trade-off between performance and efficiency than the corresponding dense ViTs.
For example, for the ViT-Tiny model, our Mobile V-MoE outperforms its dense counterpart by 3.39\% on ImageNet-1k.
For an even smaller ViT variant with only 54M FLOPs inference cost, our MoE achieves an improvement of 4.66\%.
\end{abstract}

\vspace{-3mm}
\section{Introduction}
\begin{figure}
    \centering
    \includegraphics[width=\linewidth]{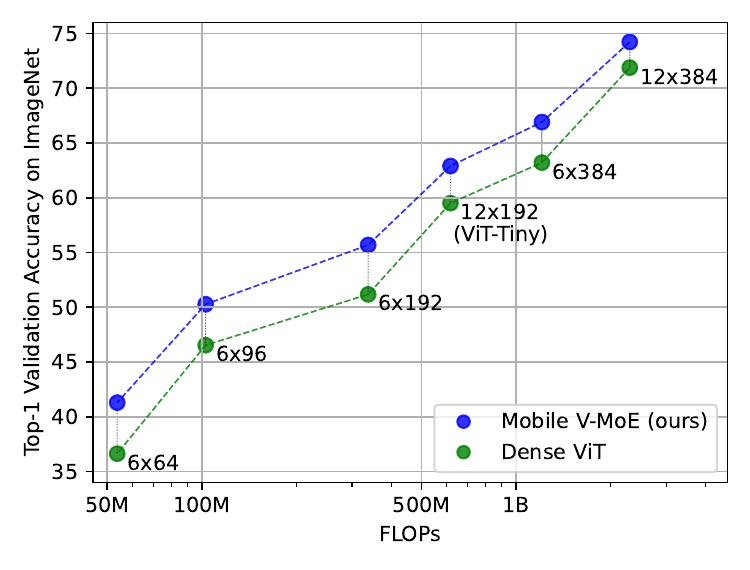}
    \caption{\textbf{Accuracy vs.~FLOPs} for ViTs of different sizes. Labels (e.g.~12$\times$192, which is ViT-Tiny) refer to the number of ViT layers (e.g.~12) and the hidden embedding dimension (e.g.~192). The sparse MoEs outperform their corresponding dense baselines across different model scales. \cref{tab:results} lists all numerical results.
    }
    \label{fig:in1k}
\end{figure}

\begin{figure*}[ht]
    \centering
    \includegraphics[width=\linewidth]{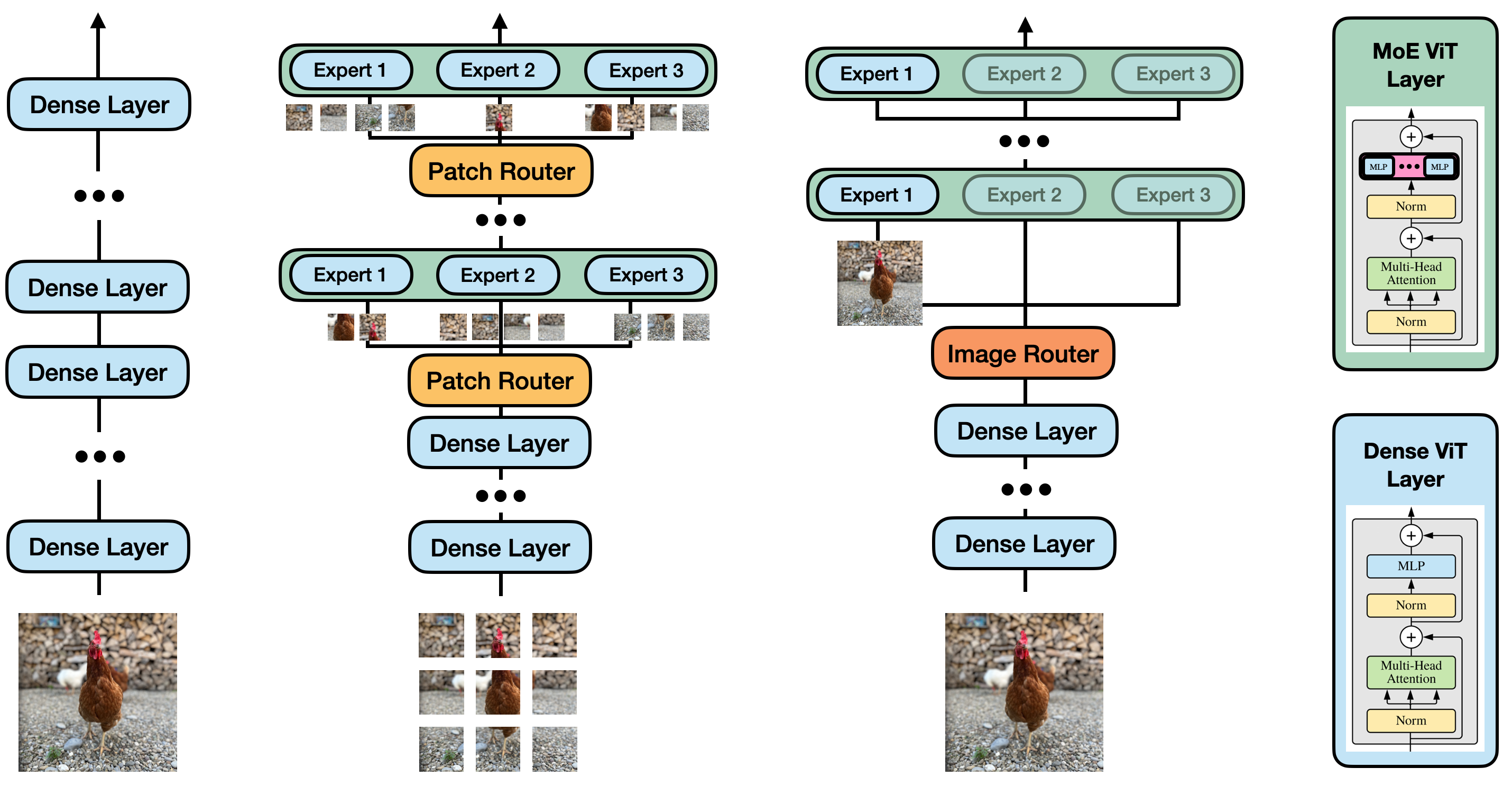}
    \begin{minipage}[t]{.13\linewidth}
        \centering
        \subcaption{Dense ViT}\label{fig:dense_vit} 
    \end{minipage}%
    \begin{minipage}[t]{.39\linewidth}
        \centering
        \subcaption{Regular V-MoE}\label{fig:regular_vmoe} 
    \end{minipage}%
    \begin{minipage}[t]{.31\linewidth}
        \centering
        \subcaption{Mobile V-MoE}\label{fig:mobile_vmoe}
    \end{minipage}%
    \begin{minipage}[t]{.20\linewidth}
        \centering
        \subcaption{Layer types}\label{fig:layers}
    \end{minipage}
    \caption{\textbf{Model architectures.} (a) The dense ViT baseline model uses dense ViT layers throughout. (b) Regular sparse V-MoE with layer-wise per-patch routers. (c) Our proposed sparse Mobile V-MoE design with a single per-image router. In both (b) and (c), dense ViT layers are followed by MoE-ViT layers (here, $k=1$ out of $E=3$ experts are activated per input). (d) In contrast to dense ViT layers \cite{vaswani2017attention}, MoE-ViT layers have a separate MLP per expert (preceded by a router) while all other parts of the layer are shared across all experts \cite{riquelme2021scaling}.}
    \label{fig:models}
\end{figure*}

The trade-off between performance and efficiency of neural networks (NNs) remains a challenge, especially in settings where computational resources are limited.
Recently, sparsely-gated Mixture-of-Experts models (sparse MoEs) have gained popularity as they provide a promising solution to this problem by enabling the decoupling of model size from inference efficiency \cite{fedus2022review}.
MoEs are NNs that are partitioned into ``experts", which are trained jointly with a router to specialize on subsets of the data. In MoEs, each input is processed by only a small subset of model parameters (aka \emph{conditional computation}).
In contrast, traditional dense models activate all parameters for each input.

Sparse MoEs were popularized in deep learning by \cite{shazeer2017outrageously}, which introduced sparse MoE-layers as drop-in replacements for standard NN layers.
Most recent MoEs are based on the Transformer~\cite{vaswani2017attention}, which processes individual input tokens; in accordance, recent MoEs also route individual input tokens to experts, i.e., image patches in the case of Vision Transformers (ViTs) \cite{riquelme2021scaling,dosovitskiy2020image} (see \cref{fig:regular_vmoe}).
Conditional computation as implemented by sparse MoEs has enabled the training of Transformers of unprecedented size \cite{fedus2021switch}.
As a result, MoEs have achieved impressive successes across various domains including language \cite{lepikhin2020gshard,fedus2021switch}, vision \cite{riquelme2021scaling}, speech \cite{you2021speechmoe} and multi-modal learning \cite{mustafa2022multimodal}, and currently hold state-of-the-art results on many benchmarks~\cite{zoph2022designing}.

The ability to increase model capacity while keeping inference cost low is also appealing for resource-constrained vision problems.
While Transformers are getting increasingly established as the de-facto standard architecture for large-scale visual modeling \cite{dosovitskiy2020image,riquelme2021scaling}, virtually all mobile-friendly models still leverage convolutions due to their efficiency \cite{howard2017mobilenets,sandler2018mobilenetv2,howard2019searching,chen2022mobile,mehta2021mobilevit,vasu2022improved}.
As such, conditional computation could potentially enable attention-based models to reduce the gap to convolutional models in the small-scale regime.
However, Transformer-based MoEs have not yet been explored for resource-constrained settings; this might be due to two main weaknesses of recently-popularized MoEs \cite{shazeer2017outrageously}.

Firstly, while per-token routing increases the flexibility to learn an optimal computation path through the model, it makes inference inefficient, as many (or even all) experts need to be loaded for a single input image.
Secondly, recent MoEs train the routers jointly with the rest or the model in an end-to-end fashion. To avoid collapse to just a few experts while ignoring all others, one needs to use load balancing mechanisms \cite{fedus2022review} such as dedicated auxiliary losses \cite{shazeer2017outrageously}. However, the resulting complex optimization objectives often lead to training instabilities / divergence \cite{lepikhin2020gshard,fedus2021switch,zoph2022designing,mustafa2022multimodal}.

In this work, we investigate the potential of sparse MoEs to scale-down ViTs for resource-constrained vision applications via an MoE design and training procedure that addresses the aforementioned issues.
Our contributions are:
\begin{enumerate}
    \item We propose a simplified, mobile-friendly sparse MoE design in which a single router assigns entire images (rather than image patches) to the experts (see \cref{fig:mobile_vmoe}).
    \item We develop a simple yet robust training procedure in which expert imbalance is avoided by leveraging semantic super-classes to guide the router training.
    \item We empirically show that our proposed sparse MoE approach allows us to scale-down ViT models by improving their performance vs.~efficiency trade-off.
\end{enumerate}

\section{Scaling down ViTs via sparse MoEs}
\subsection{Conditional computation with sparse MoEs}
An MoE implements conditional computation by activating different subsets of a NN (so-called experts) for different inputs. We consider an MoE layer with $E$ experts as
\begin{equation}
    \text{MoE}(\x) = \sum_{i=1}^E g(\x)_i e_i(\x),
\end{equation}
where $\x \in \mathbb{R}^D$ is the input to the layer, $e_i: \mathbb{R}^D \rightarrow \mathbb{R}^D$ is the function computed by expert $i$, and $g: \mathbb{R}^D \rightarrow \mathbb{R}^E$ is the routing function which computes an input-dependent weight for each expert \cite{shazeer2017outrageously}.
In a ViT-based MoE, each expert $e_i$ is parameterized by a separate multi-layer perceptron (MLP) within the ViT layer, while the other parts are shared across experts (see \cref{fig:layers}).
We use the routing function
\begin{equation}
\label{eq:router}
    g(\x) = \text{TOP}_k(\text{softmax}(\mathbf{W}\x)),
\end{equation}
where the operation $\text{TOP}_k(\x)$ sets all elements of $\x$ to zero except those with the $k$ largest values \cite{riquelme2021scaling}.
In a sparse MoE, we have $k \ll E$, s.t.~we only need to load and compute the $k$ experts with the largest routing weights. This allows us to scale-up the overall model capacity (determined by $E$) without increasing the inference cost (determined by $k$).

\subsection{Efficient and robust MoEs for small-scale ViTs}
\textbf{Per-image routing.}
Recent large-scale sparse MoEs use per-patch routing (i.e.~the inputs $\x$ are individual image patches). This generally requires a larger number of experts to be activated for each image.
For example, \cite{riquelme2021scaling} show that in their MoE with per-patch routing, ``most images use --on aggregate by pooling over all their patches-- most of the experts" \cite[Appendix E.3]{riquelme2021scaling}. Thus, per-patch routing can increase the computational and memory overhead of the routing mechanism and reduce the overall model efficiency.
We instead propose to use per-image routing (i.e., the inputs $\x$ are entire images) to reduce the number of activated experts per image, as also done in early works on MoEs \cite{jacobs1991adaptive,jordan1994hierarchical}.

\textbf{Super-class-based routing.}
Previous works on sparse MoEs jointly train the router end-to-end together with the experts and the dense ViT backbone, to allow the model to learn the optimal assignment from inputs to experts based on the data \cite{riquelme2021scaling}. While learning the optimal routing mechanism from scratch can result in improved performance, it often leads to training instabilities and expert collapse, where most inputs are routed to only a small subset of the experts, while all other experts get neglected during training \cite{fedus2022review}. Thus, an additional auxiliary loss is typically required to ensure load-balancing between the different experts, which can increase the complexity of the training process \cite{fedus2022review}.

In contrast, we propose to group the classes of the dataset into super-classes and explictly train the router to make each expert specialize on one super-class.
To this end, we add an additional cross-entropy loss $\mathcal{L}_g$ between the router output $g(\x)$ in \cref{eq:router} and the ground truth super-class labels to the regular classification loss $\mathcal{L}_C$ to obtain the overall weighted loss $\mathcal{L} = \mathcal{L}_C + \lambda \mathcal{L}_g$ (we use $\lambda=0.3$ in our experiments, which we found to work well).
Such a super-class division is often readily provided with the dataset (e.g.~for CIFAR-10/100 or MS-COCO). 
If a dataset does not come with a super-class division, we can easily obtain one as follows: 1) we first train a dense baseline model on the dataset; 2) we then compute the model's confusion matrix over a held-out validation set; 3) we finally construct a confusion graph from the confusion matrix and apply a graph clustering algorithm to obtain the super-class division \cite{jin2017confusion}.
This approach encourages the super-classes to contain semantically similar images that the model often confuses.
Intuitively, by allowing the different MoE experts to specialize on the different semantic data clusters, performance on the highly-confused classes should be improved.
We use this approach in our experiments on ImageNet-1k, computing the confusion matrix via a dense ViT-S/16 model.
The resulting super-class division for $E=10$ experts is shown in \cref{tab:super_classes}; the super-classes contain semantically related classes.

\begin{table}
    \centering
    \begin{tabular}{lcc}
    \toprule
    \textbf{ID} & \textbf{Classes} & \textbf{Super-class}\\
    \midrule
    0 & boxer, pug, Rottweiler & dogs\\
    1 & orangutan, weasel, panda & other mammals\\
    2 & toucan, flamingo, ostrich & birds\\
    3 & eel, scorpion, hammerhead & other animals\\
    4 & minivan, ambulance, taxi & land vehicles\\
    5 & submarine, canoe, pirate & sea vehicles\\
    6 & guacamole, hotdog, banana & food\\
    7 & backpack, pyjama, kimono & clothes\\
    8 & monitor, iPod, photocopier & tech devices\\
    9 & xylophone, harp, trumpet & instruments\\
    \bottomrule
    \end{tabular}
    \caption{\textbf{Super-class division for $E=10$}. For each super-class, we list three randomly chosen class names (which turn out to be semantically related) together with a possible super-class name.}
    \label{tab:super_classes}
\end{table}

\section{Experiments}
We now present empirical results on the standard ImageNet-1k classification benchmark \cite{russakovsky2015imagenet}.
We train all models from scratch on the ImageNet-1k training set of 1.28M images, and then evaluate their top-1 accuracy on the held-out validation set of 50K images.
In \cref{subsec:in1k}, we first evaluate our proposed sparse Mobile V-MoE across a range of model scales and show that they achieve better performance vs.~efficiency trade-offs than the respective dense ViT baselines.
In \cref{subsec:ablations}, we then conduct several ablation studies to get a better understanding of the properties of our proposed sparse MoE model design and training procedure.
\subsection{Accuracy vs.~efficiency across ViT scales}
\label{subsec:in1k}
\begin{figure*}
    \small
    \begin{minipage}{.4\textwidth}
     \begin{subfigure}[b]{\linewidth}
        \centering
        \begin{tabular}{lrrrr}
        \toprule
        \textbf{Model} & \textbf{FLOPs} & \multicolumn{3}{c}{\textbf{Top-1 Accuracy}}\\
        \cmidrule(lr){3-5}
        & & \textbf{Dense} & \textbf{MoE} & $\mathbf{\Delta}$\\
        \midrule
            12$\times$384  & 2297M & 71.88 & \textbf{74.23} & {\color{teal}\textbf{+2.35}}\\
            9$\times$384 & 1752M & 69.94 & \textbf{72.47} & {\color{teal}\textbf{+2.53}}\\
            6$\times$384 & 1207M & 63.21 & \textbf{66.91} & {\color{teal}\textbf{+3.70}}\\
            12$\times$192\footnotemark{}& 618M & 59.51 & \textbf{62.90} & {\color{teal}\textbf{+3.39}}\\
            9$\times$192 & 478M & 56.50 & \textbf{59.52} & {\color{teal}\textbf{+3.02}}\\
            6$\times$192 & 338M & 51.18 & \textbf{55.69} & {\color{teal}\textbf{+4.51}}\\
            12$\times$96 & 176M & 53.79 & \textbf{55.39} & {\color{teal}\textbf{+1.60}}\\
            9$\times$96& 140M & 51.27 & \textbf{52.99} & {\color{teal}\textbf{+1.72}}\\
            6$\times$96& 103M & 46.54 & \textbf{50.28} & {\color{teal}\textbf{+3.74}}\\
            12$\times$64 & 88M & 42.90 & \textbf{46.07} & {\color{teal}\textbf{+3.17}}\\
            9$\times$64& 71M & 40.46 & \textbf{43.95} & {\color{teal}\textbf{+3.49}}\\
            6$\times$64& 54M & 36.64 & \textbf{41.30} & {\color{teal}\textbf{+4.66}}\\
        \bottomrule
        \end{tabular}
        \caption{\textbf{Accuracy vs.~efficiency across ViT scales.}}
        \label{tab:results}
     \end{subfigure}
    \end{minipage}%
    \begin{minipage}{.25\textwidth}
     \begin{subfigure}[b]{\linewidth}
        \centering
        \begin{tabular}{lrrr}
        \toprule
        $\mathbf{E}$ & \textbf{Router} & \textbf{MoE} & $\mathbf{\Delta}$\\
        \midrule
        5 & 86.43 & 72.33 & {\color{teal} +1.64}\\
        7 & 87.43 & 73.13 & {\color{teal} +2.44}\\
        \textbf{10} & 87.12 & 73.52 & {\color{teal} +2.83}\\
        15 & 84.16 & 73.10 & {\color{teal} +2.41}\\
        20 & 84.08 & 73.36 & {\color{teal} +2.67}\\
        \bottomrule
        \end{tabular}
        \caption{\textbf{Total number of experts $E$}.}
        \label{tab:E}
        \vspace{0.09cm}
     \end{subfigure}
     \hfill
     \begin{subfigure}[b]{\linewidth}
        \centering
        \begin{tabular}{lrrr}
        \toprule
        $\mathbf{L}$ & \textbf{Router} & \textbf{MoE} & $\mathbf{\Delta}$\\
        \midrule
        1 & 90.17 & 72.14 & {\color{teal} +1.45}\\
        \textbf{2} & 87.12 & 73.52 & {\color{teal} +2.83}\\
        4 & 82.12 & 71.67 & {\color{teal} +0.98}\\
        6 & 77.70 & 70.07 & {\color{red} -0.62}\\
        8 & 72.09 & 64.47 & {\color{red} -6.22}\\
        \bottomrule
        \end{tabular}
        \caption{\textbf{Number of MoE layers $L$}.}
        \label{tab:L}
     \end{subfigure}
    \end{minipage}%
    \begin{minipage}{.35\textwidth}
     \begin{subfigure}[b]{\linewidth}
        \centering
        \begin{tabular}{lcrrr}
        \toprule
        $\mathbf{k}$ & \textbf{FLOPs} & \textbf{Dense} & \textbf{MoE} & $\mathbf{\Delta}$\\
        \midrule
        \textbf{1} & 2534M & 70.79 & 73.44 & {\color{teal} +2.65}\\
        2 & 2769M & 71.70 & 74.42 & {\color{teal} +2.72}\\
        3 & 3005M & 73.58 & 74.44 & {\color{teal} +0.86}\\
        5 & 3476M & 74.87 & 74.32 & {\color{red} -0.55}\\
        10 & 4653M & 75.10 & 74.37 & {\color{red} -0.73}\\
        \bottomrule
        \end{tabular}
        \caption{\textbf{Number of experts $k$ per image}.}
        \label{tab:num_experts}
        \vspace{0.09cm}
     \end{subfigure}
     \begin{subfigure}[b]{\linewidth}
        \centering
        \begin{tabular}{lcrr}
        \toprule
        \textbf{Routing} & \textbf{Input} & \textbf{Acc.} & $\mathbf{\Delta}$\\
        \midrule
        Dense & N/A & 71.70 &\\
        \textbf{Super-class} & image & 74.42 & {\color{teal} +2.72}\\
        Rand. class & image & 69.22 & {\color{red} -2.48}\\
        End-to-end & image & 73.57 & {\color{teal} +1.87}\\
        End-to-end & token & 74.85 & {\color{teal} +3.15}\\
        \bottomrule
        \end{tabular}
        \caption{\textbf{Routing strategies}.}
        \label{tab:routing}
     \end{subfigure}
    \end{minipage}
    \caption{\textbf{Empirical results.} (a) Our Mobile V-MoEs outperform the respective dense ViTs across model scales. Model names (e.g.~12$\times$192) refer to the number of layers (12) and the embedding size (192). (b-e) Ablation studies using DeiT-Ti/16 \cite{touvron2021training}, with $k=1$, $E=10$, $L=2$ by default. Best performance vs.~efficiency trade-off is achieved with (b) $E=10$ experts total, (c) $L=2$ MoE layers (out of 12 layers total), (d) $k=1$ or $k=2$ experts activated per image, (e) our semantic super-class routing; the settings used in (a) are bolded.}
    \label{fig:results}
\end{figure*}

We consider ViT models (both MoEs and corresponding dense baselines) of different sizes by scaling the total number of layers (we use 12, 9 or 6) and the hidden embedding size (we use 384, 192, 96 or 64).
The number of multi-head self-attention heads is (6, 3, 3, 2) for the different hidden embedding sizes.
The embedding size of the MLP is $4 \times$ the hidden embedding size, as is common practice.
We use $E=10$ experts in total for the MoE, out of which $k=1$ is activated per input image.
Our MoEs comprise of $L=2$ MoE-ViT layers preceded by (10, 7 or 4) dense ViT layers (see \cref{fig:mobile_vmoe}).
We use a patch size of $32\times32$ for all models.
This is because the the patch size effectively controls the trade-off between FLOPs and number of model parameters: as we aim to optimize for FLOPs, a larger patch size (resulting in a fewer number of patches) is beneficial.
We also tried using a smaller patch size of $16\times16$, where the result trends were basically the same (but where the number of FLOPs was higher relative to the model capacity and thus accuracy).
For the ViTs with hidden sizes 384 and 192, we use the DeiT training recipe \cite{touvron2021training}, while for hidden sizes 96 and 64, we use the standard ViT training recipe \cite{dosovitskiy2020image} to avoid underfitting.
\cref{fig:in1k,tab:results} compare top-1 validation accuracy vs.~FLOPs.
Our Mobile V-MoEs outperform the corresponding dense ViT baselines across all model sizes.

\footnotetext{This corresponds to the ViT-Tiny model \cite{touvron2021training} with patch size $32\times32$.}

\subsection{Ablation studies}
\label{subsec:ablations}
We train DeiT-Tiny \cite{touvron2021training} (12 layers total, 192 embedding size, $16\times16$ patch size) with $k=1$ out of $E=10$ experts per input, and with $L=2$ MoE layers (unless noted otherwise); the dense ViT baseline achieves 70.79\% accuracy.

\textbf{Total number of experts $E$.}
We consider different widths of the MoE, i.e., different numbers of experts $E$ (and thus super-classes), ranging between $E=5$ and $E=20$.
We report both the accuracy of the entire MoE model (i.e., on the 1,000-way classification task), as well as the accuracy of the router (i.e., on the $E$-way super-classification task).
\cref{tab:E} shows that overall performance improves until $E=10$, from which point onwards it stagnates. The router accuracy also drops beyond $E=10$ due to the increased difficulty of the $E$-way super-classification problem.

\textbf{Number of MoE layers $L$.}
We consider different depths of the MoE, i.e., different numbers of MoE layers $L$, ranging between $L=1$ and $L=8$ (out of 12 ViT layers in total).
We again report both the full MoE and router accuracies.
\cref{tab:L} shows that overall performance peaks at $L=2$, and rapidly decreases for larger $L$. This is due to the router accuracy, which declines with increasing $L$ as the router gets less information (from the $12-L$ ViT layers).

\textbf{Number of experts $k$ per image.}
We vary the number of experts $k$ activated per image.
We compare against dense baselines that use an MLP with hidden dimension scaled by $k$ to match the MoE's inference FLOPs.
\cref{tab:num_experts} shows that $k=1$ and $k=2$ perform best (relative to the dense baseline), with decreasing performance delta for larger~$k$. 

\textbf{Routing strategies.}
We compare our proposed semantic super-class per-image routing vs.~end-to-end-learned routing (both per-image and per-token) and a baseline with random super-classes (for $k$=2).
\cref{tab:routing} shows that our method (\cref{fig:mobile_vmoe}) is better, except for learned per-token routing (as in the regular V-MoE \cite{riquelme2021scaling}, \cref{fig:regular_vmoe}), which however needs to activate many more experts and thus model parameters for each input image (up to 11.05M, vs.~6.31M for ours).

\section{Conclusions and future work}
We showed that sparse MoEs can improve the performance vs.~efficiency trade-off compared to dense ViTs, in an attempt to make ViTs more amenable to resource-constrained applications.
In the future, we aim to apply our MoE design to models that are more mobile-friendly than ViTs, e.g., light-weight CNNs such as MobileNets \cite{howard2017mobilenets,sandler2018mobilenetv2,howard2019searching} or ViT-CNN hybrids \cite{chen2022mobile,mehta2021mobilevit,vasu2022improved}.
We also aim to consider other vision tasks, e.g., object detection.
Finally, we aim to get actual on-device latency measurements for all models.

\clearpage
{\small
\bibliographystyle{ieee_fullname}
\bibliography{egbib}
}

\end{document}